\documentclass[a4paper, 10pt, conference]{IEEEtran}
\IEEEoverridecommandlockouts
\usepackage{cite}
\usepackage{amsmath,amssymb,amsfonts}
\usepackage{algorithmic}
\usepackage{graphicx}
\usepackage{textcomp}
\usepackage{tabularx}
\usepackage{amsmath}

\usepackage{float}
\usepackage{amsmath}
\DeclareMathOperator{\atantwo}{atan2}

\usepackage{xcolor}
\def\BibTeX{{\rm B\kern-.05em{\sc i\kern-.025em b}\kern-.08em
 T\kern-.1667em\lower.7ex\hbox{E}\kern-.125emX}}
\begin{document}
 
\title{HyperGuider: Virtual Reality Framework for Interactive Path Planning of Quadruped Robot in Cluttered and Multi-Terrain Environments
}

\makeatletter
\newcommand{\linebreakand}{%
 \end{@IEEEauthorhalign}
 \hfill\mbox{}\par
 \mbox{}\hfill\begin{@IEEEauthorhalign}
}
\makeatother

\author{
 \IEEEauthorblockN{Ildar Babataev}
 \IEEEauthorblockA{\textit{Intelligent Space Robotics Laboratory} \\
 \textit{Skoltech}\\
 Moscow, Russian Federation \\
 ildar.babataev@skoltech.ru}
\and
 \IEEEauthorblockN{Aleksey Fedoseev}
 \IEEEauthorblockA{\textit{Intelligent Space Robotics Laboratory} \\
 \textit{Skoltech}\\
 Moscow, Russian Federation \\
 aleksey.fedoseev@skoltech.ru}
\and
 \IEEEauthorblockN{Nipun Weerakkodi}
 \IEEEauthorblockA{\textit{Intelligent Space Robotics Laboratory} \\
 \textit{Skoltech}\\
 Moscow, Russian Federation \\
 nipun.weerakkodi@skoltech.ru}
\linebreakand
 \IEEEauthorblockN{Elena Nazarova}
 \IEEEauthorblockA{\textit{Intelligent Space Robotics Laboratory} \\
 \textit{Skoltech}\\
 Moscow, Russian Federation \\
 elena.nazarova@skoltech.ru}
\and
 \IEEEauthorblockN{Dzmitry Tsetserukou}
 \IEEEauthorblockA{\textit{Intelligent Space Robotics Laboratory} \\
 \textit{Skoltech}\\
 Moscow, Russian Federation \\
 d.tsetserukou@skoltech.ru}
}
\maketitle
\begin{abstract}
Quadruped platforms have become an active topic of research due to their high mobility and traversability in rough terrain. However, it is highly challenging to determine whether the clattered environment could be passed by the robot and how exactly its path should be calculated. Moreover, the calculated path may pass through areas with dynamic objects or environments that are dangerous for the robot or people around. Therefore, we propose a novel conceptual approach of teaching quadruped robots navigation through user-guided path planning in virtual reality (VR). 
Our system contains both global and local path planners, allowing robot to generate path through iterations of learning. The VR interface allows user to interact with environment and to assist quadruped robot in challenging scenarios. 

The results of comparison experiments show that cooperation between human and path planning algorithms can increase the computational speed of the algorithm by 35.58\% in average, and non-critically increasing of the path length (average of 6.66\%) in test scenario. Additionally, users described VR interface as not requiring physical demand (2.3 out of 10) and highly evaluated their performance (7.1 out of 10). The ability to find a less optimal but safer path remains in demand for the task of navigating in a cluttered and unstructured environment.

\end{abstract}


\section{Introduction}
Today, mobile robots are widely applied in exploration, mapping, and rescue missions in cluttered and unsafe environments. Researchers actively explore the ability of mobile robots to navigate on rough terrain to expand the scope of their application. For example, robot motion planning in dynamic uncertain environments was explored by Toit et al. \cite{Toit_2010}. Path planning on rough terrain with multi-objective particle swarm optimization was developed by Wang et al. \cite{Wang_2018} for non-holonomic robots. Dijkstra's algorithm with cost function depending on terrain traversability and robot power consumption was suggested by Santos et al. for path planning on rough terrain \cite{Santos_2019}. 
\begin{figure}[!h]
 \includegraphics[width=1.0\linewidth]{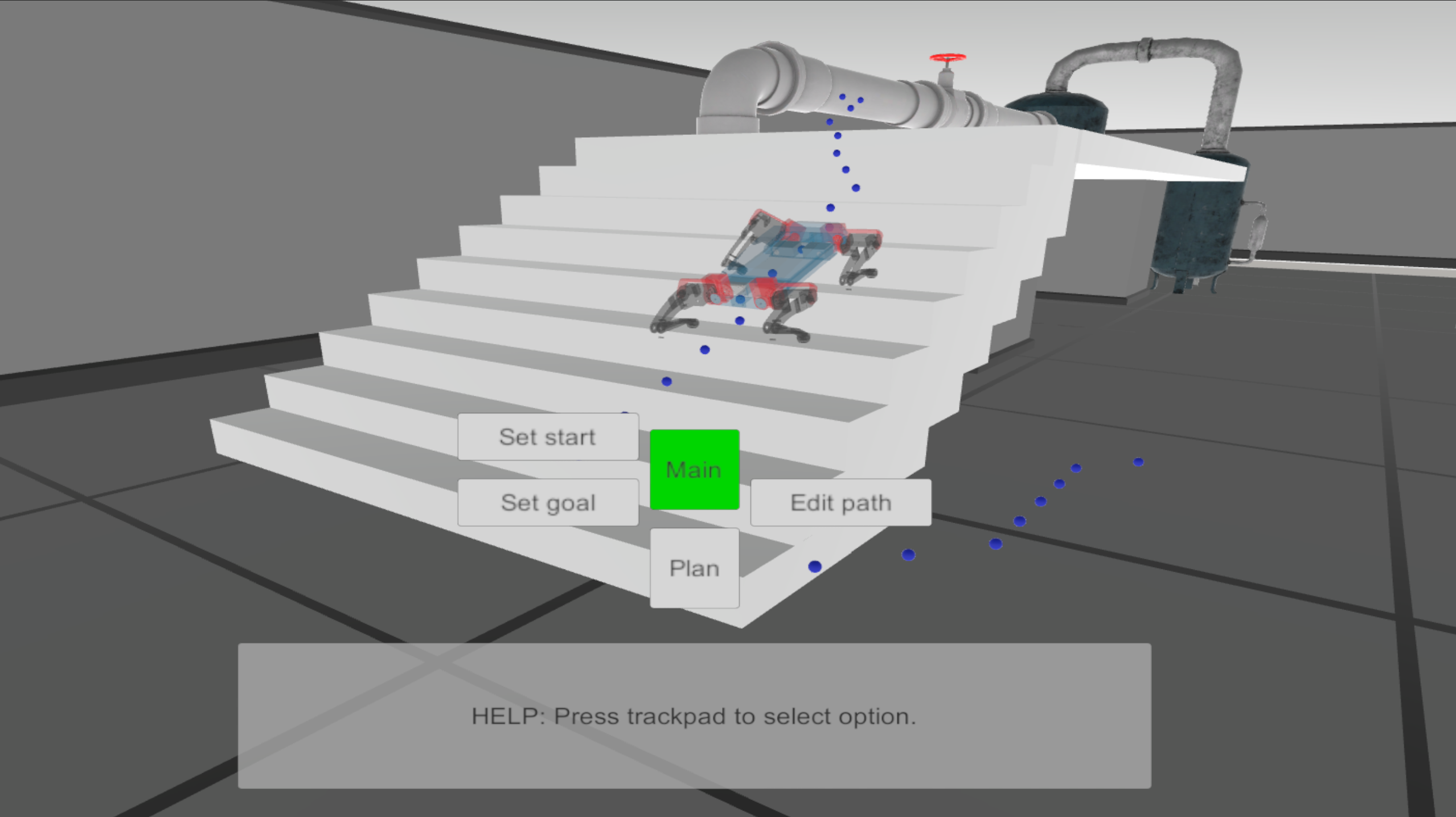}
 \caption{ HyperGuider virtual reality user interface (VRUI) module. }
 \label{fig:main}
\vspace{-1.5em}
\end{figure}

A kinodynamic discrete planning that improves the efficiency of the sampling-based motion algorithms was proposed by Samaniego et al. \cite{Samaniego_2020} for navigation in cluttered environment.

Among all types of robots, quadruped platforms have a high potential due to their ability to explore the terrain with a higher number of obstacles compared to wheeled robots \cite{Lee_2020}. For example, Hutter et al. \cite{Hutter_2017} presented a highly mobile quadruped platform developed for autonomous operation in harsh environments. Barasuol et al. \cite{Barasuol_2013} introduced a reactive controller for quadrupedal locomotion on challenging terrains. Self-organized locomotion with control based on neural networks was explored by Sun et al. \cite{Sun_2018}, where researchers demonstrated a successful simulation of a legged robot in presence of low-height obstacles. 

However, there is high technological complexity in developing and operating such robots, which involves advanced mechatronics, robotics, control theory, etc. \cite{Dang_2019}. The problem of path planning presents one of the critical challenges to be addressed in order to ensure reliable navigation of a four-legged robot. Several recent works were focused on locomotion and gait planning. For example, Aeini et al. \cite{Aeini_2022} presented a straightforward approach for trot gait planning at the desired speed, while preserving the stability of the robot. Hu et al. \cite{Hu_2014} designed a controller to maintain a stable walk of a quadruped robot with a trotting gait for walking designed to achieve a higher speed on irregular terrain. In addition, a number of projects are devoted to the recognition and classification of the objects and surfaces for path planning. Li et al. \cite{Li_2020} proposed a terrain classification and prediction method, which improves computational efficiency and reacts to the overall terrain of the path.

Despite the ability of these approaches to improve autonomous path planning in presence of obstacles, industrial and hazardous environments include a high number of unstable or moving objects and complex surfaces. This makes the task of autonomous navigation highly difficult for a robot, while a human can solve it by relying on their cognitive capabilities. 
In this paper, we propose a novel control system for quadruped mobile robot path planning based on virtual reality (VR). Our system contains a user interface for interacting with a path planning module and a planning algorithm that learns to navigate in a virtual environment, adjusting taking into account operator commands.

\section{Related Work}
The system we propose is closely related to the concepts of teleoperation and remote control, which have been incredibly enhanced by an immersive virtual environment VR tools are being actively suggested for teleoperation due to their potential of reducing mental demand and limiting the workload of sensory modalities leading to higher performance of operators \cite{Ostanin_2018}. Several prior works explored remote control via virtual interfaces. For example, Bo et al. \cite{Bo_2014} has introduced a teleoperation system in VR which enhances the operator ability via a preliminary simulation of the task in a virtual scene. However, modern systems achieved higher level of perception and navigation, as a result operator do not need to control all movements of the mobile robot. Mulun et al. \cite{Mulun_2020} suggested to use Mixed Reality (MR) for interactive path planning in which operator can create virtual obstacles on the way to keep safe path. System already uses VFH* path planning algorithm to find path for wheeled omnidirectional mobile robot.

Current path planning methods include classical A*-based and D*-based algorithms, algorithms based on artificial potential fields \cite{Csiszar_2012}, etc. For example, Li et al \cite{Li_2019} introduced ADFA* path planning algorithm which uses the dilation factor based on DFA* and provides a path search result related to the time limit. However, the proposed algorithm operates only with discrete linear coordinates of the robot and can only be considered within the global path planning, while local path planning is required to take into account quadruped kinematics. Yufei et al. \cite{Yufei_2020} introduced a path planning module based on the Dijkstra algorithm with Timed Elastic Band smoothing for global terrain navigation and obstacle avoidance algorithm for local map based on potential fields. 
Model-based reinforcement learning approaches for path planning were significantly improved in the last years. For example, Kulathunga \cite{Kulathunga_2021} developed a hybrid approach that combines Monte Carlo tree search and RL-based approach, to achieve higher accuracy than Proximal policy optimization, deep Q-network, and uniform tree search.

Despite the advances in path planning algorithms, the effective combination of global and local planning methods to traverse the unstructured terrains remains challenging and time-consuming, thus, requiring further exploration.

\section{System Overview}

Our developed system contains three main modules: Virtual Reality user interface (VRUI), Mapping module, and Navigation module. All modules are implemented with the C\# programming language in the Unity Engine framework. Fig.~\ref{fig:short_SO} shows the communication between these modules.

\begin{figure}[htbp]
\centerline{\includegraphics{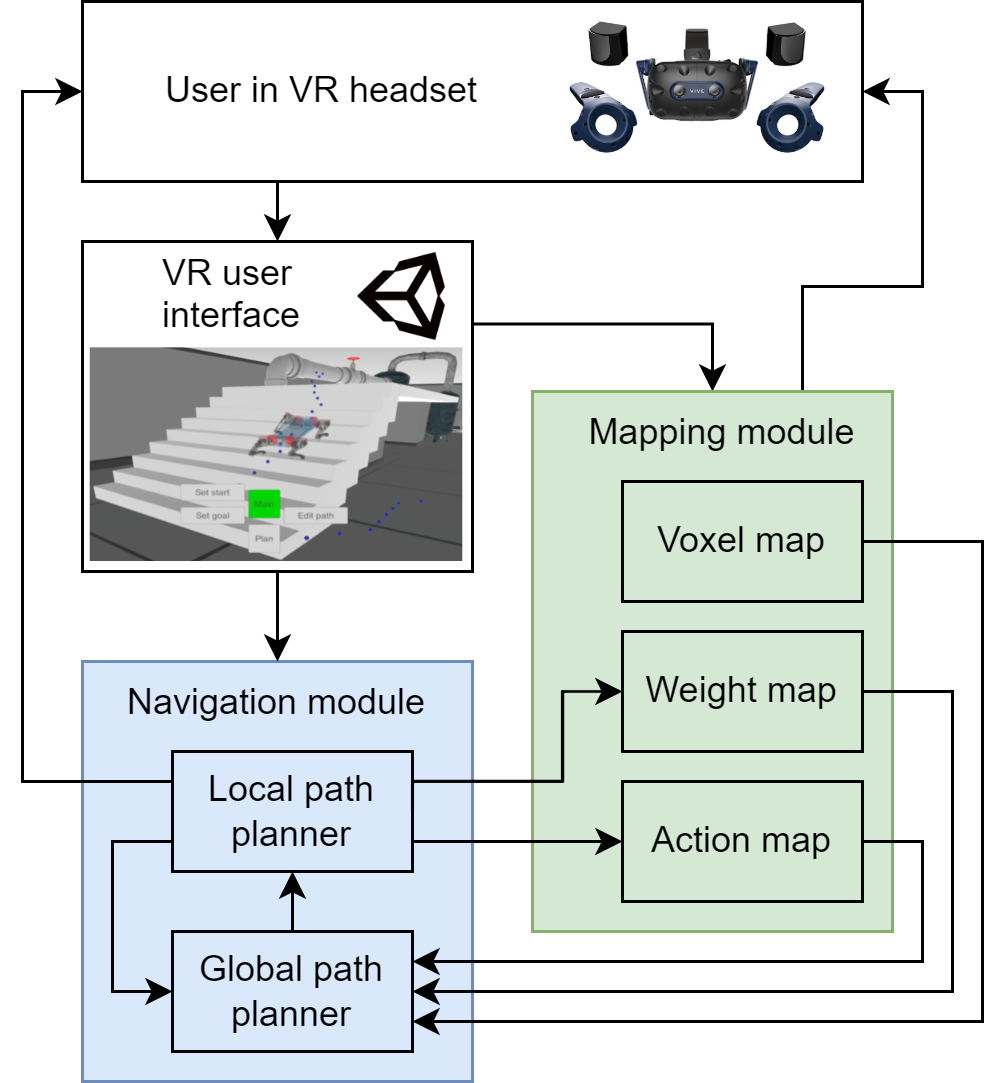}}
\caption{Overview of the HyperGuider system with Navigation, Mapping and VR user interface modules.}
\label{fig:short_SO}
\end{figure}

VRUI is implemented for immersive interaction between the operator and the path planning algorithm. The interface is developed for the HTC Vive Pro headset and allows operators to generate virtual obstacles and edit poses of a global path with controllers. In addition, the operator can set the start and goal position and orientation of the quadruped robot. 

The Mapping module includes three map types: voxel map, positional weight map, and action weight map. The voxel map is an array of cell coordinates occupied by objects of the environment Fig.~\ref{fig:VRwithVoxelMap}. The positional weight map is a dictionary of discrete states with additional cell costs $w(q_D)$. Action weight map represents the additional cost of action $a$ from the given discrete state $q_D$. The user interacts with the planning process using a weight map by creating virtual obstacles.

\begin{figure}[htbp]
\centerline{\includegraphics[width=0.45\textwidth]{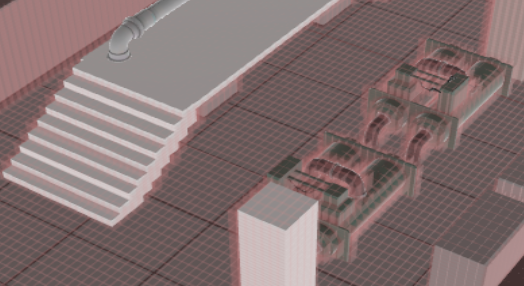}}
\caption{Virtual environment combined with voxel map. The red voxels are occupied with obstacles.}
\label{fig:VRwithVoxelMap}
\end{figure}

The Navigation module is the core of functionality, including global and local path planning algorithms. The global path planner uses information about the discrete environment from the Mapping module. It takes into account the voxel map, positional, and action weight maps to calculate the global discrete plan $G_D$. The local path planning algorithm receives the $G_D$ and searches for the sequence of obstacle-free states $S$. It also contains the digital twin of our quadruped robot HyperDog.


The algorithm is able to find invalid states and modify the positional and the action weight maps for a future iteration of the global algorithm. This module solves the inverse kinematics problem for the quadruped robot to define its valid states in space.

\section{Path Planning}

\subsection{Global Path Algorithm}
The global path planning module is developed to find a sequence of obstacle-free discrete coordinates $q_D$ in existed static voxel map. The algorithm based on A* was developed to find a global path $G_D$, which is then processed by the local path planning module. In addition, the discrete global path can be easily visualized in the VR environment.

Global planning algorithm is developed under the following assumptions:
\begin{itemize}
\item Configuration space $C$ is presented by three dimensional manifold of discrete linear coordinates $q_D = [x_D, y_D, z_D]^T$ of the trunk body.
\item Action space $A$ allows the robot to move in any direction to the nearest voxels. The default cost of each action $a \in A$ is equal to Euclidean distance. 
\item The vertical distance from the obstacle to the robot's body must not exceed the maximum standing height $h_{th}$. This value is calculated using inverse kinematics task and voxel map resolution.
\end{itemize}

Heuristic function of the global planning algorithm uses Euclidean distance between discrete state and goal state. Its calculation is given by:

\begin{equation}
h(q_D) = || g_D - q_D || + \hat{h}(q_D) \label{eq_heuristic}
\end{equation}
\noindent
where $q_D$ is the discrete state, $g_{D}$ is the goal state, $\hat{h}(q_D)$ is the additional cost. Additional cost can be defined by the local path planning module or user interaction.
The cost of the action is defined from the state by: 

\begin{equation}
f(a, q_D) = || a || + \hat{f}(a, q_D) \label{eq_action}
\end{equation}
\noindent
where $a$ is the current action, $q_D$ is the current state, $\hat{f}(a, q_D)$ is the additional cost, which is defined by local planning module.

\subsection{Local Path Algorithm}

The local planning module is a sampled-based algorithm. It searches the obstacle-free robot states with valid kinematic configuration. The result of this algorithm is a sequence of states $s$, which can be described by 18 variables:

\begin{itemize}
\item Position of the trunk body $q^p = [x, y, z]^T$.
\item Orientation of the trunk body $q^r = [\phi, \psi, \gamma]^T$.
\item Matrix 4 by 3 of the leg joint angles $\Theta$. Quadruped robot moves with four legs, each leg has three joints.
\end{itemize}

The first step of the local planning algorithm is to obtain a discrete global path $G_D$ from the global planning module. If any global solution exists, the algorithm converts $G_D$ to continuous representation $G$ and propagates from the start state $s_0$ to the first global state $g_0 \in G$. Each global state $g_i$ provides information about the next desired coordinates of the robot body and doesn't take into account desired rotations and joint angles.

The next step is to define the transformation matrix $T^{i+1}_{i}$, which includes translation and rotation parts. This matrix is given as follows:

\begin{equation}
\ T^{i+1}_{i} = 
\begin{bmatrix}
 R(\hat{\delta_r}) & \hat{\delta_p} \\
 0 & 1 \\
\end{bmatrix}
\label{eq_Transform}
\end{equation}

\noindent
where $\delta_p$ is the translation vector between the robot position $q^{p}_i$ and the state $g_i + d$. It can be found from the following equation:

\begin{equation}
\delta_p = g_i + d - q^{p}_i 
\label{eq_poses}
\end{equation}

\noindent
where $d$ is the repulsive vector. We used orientation of the current robot origin $O_i$ and orientation of the origin $O_g$ constructed by $\delta_p$ vector and global basis vector $k$ (aligned with z axis of the world) to define rotation $R$. The rotation angles between origins $O_i$ and $O_g$ are defined as a rotation vector $\delta_r$. The calculation of the rotation vector is given by:

\begin{equation}
\delta_r = 
 \begin{Bmatrix}
 \alpha = \atantwo(R_{21}, R_{11}) \\
 \beta = \atantwo(-r_{31}, \sqrt{r^{2}_{32} + r^{2}_{33}}), \\
 \gamma = \atantwo(R_{32}, R_{33}) \\
 \end{Bmatrix} \label{eq_angles}
\end{equation}

\noindent 
where $R_{11}$, $R_{21}$, $R_{32}$, $R_{33}$ are the elements of the rotation matrix $R(\delta_r)$ between origins $O_i$ and $O_g$. Both vectors $\delta_p$ and $\delta_r$ are clamped by $\delta^{max}_p$ and $\delta^{max}_r$. The resulted vectors $\hat{\delta_p}$ and $\hat{\delta_r}$ are used for a propagation to the next state $s_{i+1}$ from the current state $s_i$. 

Further steps of the local planning algorithm are dedicated to the validation of the state $s_{i+1}$. The state is considered to be valid if, firstly, there are no collisions, secondly, the inverse kinematics task is resolvable for each leg and, lastly, three out of four legs are standing on the surface. The algorithm calculates the poses of paws relative to each heap joint $r_{i+1}$. It takes a feasible relative pose from the current state $r_{i}$ and appends the heap translation vector $\delta$. Translation vector $\delta$ is calculated with known $\hat{\delta_p}$ and $\hat{\delta_r}$. 

In addition, we implemented a simple gait control algorithm. It accumulates heap transition vectors $\delta$ for each leg and selects one with the biggest accumulated value to lift it above the surface. The next step is to resolve given $r_{i+1}$ vectors with the inverse kinematics task and to check robot collisions. We accept state $s_{i+1}$ if it satisfies requirements described above. Otherwise, we calculate repulsive vector $d$ as vertical displacement along $k$. Thus, the repulsive vector is found from:

\begin{equation}
d = 
\begin{cases}
 \frac{\sum_k^{N}{(q_{i+1} - p_k + n_k)}}{N} & \text{ if } N > 0 \\ 
 -k h_{th} & \text{ if } N = 0
\end{cases}
\label{eq_repulsive} 
\end{equation}
\noindent

\noindent
where $p_k$ is the collision point and $n_k$ is the normal vector to the surface at the point of collision, $N$ is the number of estimated collisions. 

The algorithm accumulates $d$ to avoid stacking in the local minimum. It checks equality between discrete representation of state $g_i$ and $g_i + d$. If they are different, the local algorithm defines it as an invalid global planning path around $g_i$. Local planning module calculates action $a$ from $g_{i-1}$ to $g_i$ and increases the weight of this action to avoid global pose $g_i$. In addition, it appends $||d||$ to the weight of the discrete state $g_i$. Finally, it goes back to the previously achieved global pose $g_{i-1-j}$, where $j$ is the additional shift. State $g_i$ is achieved if $||\hat{\delta_p}||$ is less than $\delta^{max}_p$. The algorithm propagates to the goal state $s_g$ when the last global state is achieved. This state is counted as achieved if $||\hat{\delta_p}||$ is less than $\delta^{max}_p$ and $||\hat{\delta_r}||$ is less than $\delta^{max}_r$.

\section{Experiments}
\subsection{Path Planning without the User Interaction}
In this paper, we evaluated the influence of the operator on the navigation process. The performance of the developed path planning algorithm was firstly observed without user interaction.

\subsubsection{Procedure} We conducted an experiment with a planning algorithm that had an empty weight position and action maps at the beginning. The start and goal states were located at different heights, the way to the goal passed through the stairs with 10 steps. 
The algorithm was launched until the path length became constant. The computation time, path length, and number of requests from the local planning algorithm to the global algorithm were selected as efficiency metrics. 

\subsubsection{Results} 

The results of the experiment are presented in Table~\ref{tab:exp1}.

\begin{table}[htbp]
\caption{Performance of Path Planning Algorithm without User Interaction}
\begin{center}
\begin{tabular}{|c|c|c|c|c|c|}
\hline
\multicolumn{1}{|l|}{Iteration number}   & 1 & 2 & 3 & 4 & 5 \\ \hline
\begin{tabular}[c]{@{}c@{}}Computation\\ time, sec\end{tabular} & 27.1 & 4.5 & 4.2 & 3.9 & 3.9 \\ \hline
Path length (number of states)       & 430 & 402 & 400 & 399 & 399 \\ \hline
Request count      & 76 & 3 & 3 & 1 & 1 \\ \hline
\end{tabular}
\label{tab:exp1}
\end{center}
\end{table}

The planning algorithm was able to find a path in each iteration. It took four loops of the algorithm to find the shortest path, which length is 92.7\% from the one at the first iteration. The computation time was also reduced by 85.6\%. We can observe that the shortest path corresponds to the absence of additional requests from the local planner, meaning only one request at the beginning was required. The total time required to find the optimal path is 39.7 seconds. 

In addition, we have analyzed changes in the weights during the learning process, shown in Fig.~\ref{fig:weightsSetted}. The final global path and positional weights distribution are shown in Fig.~\ref{fig:exp1_map}.

\begin{figure}[htbp]
\begin{center}
\centerline{\includegraphics[width=0.5\textwidth]{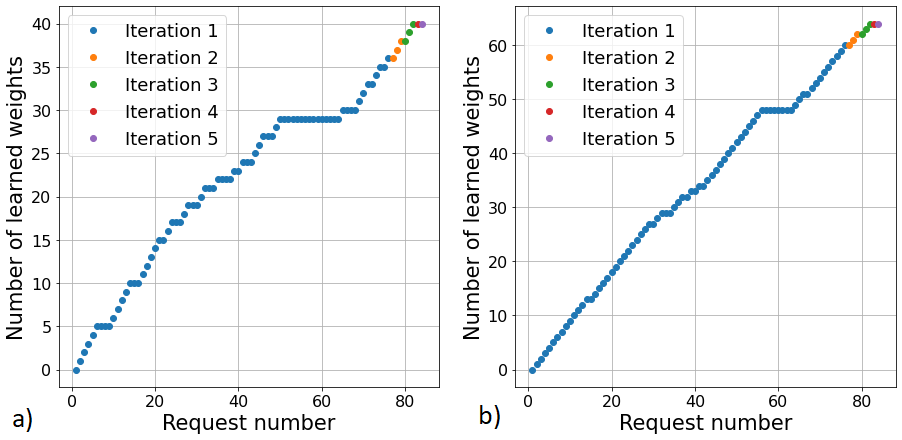}}
\caption{Dynamics of setting (a) position and (b) action weights during autonomous planning process.}
\label{fig:weightsSetted}
\end{center}
\vspace{-0.5em}
\end{figure}

\begin{figure}[htbp]
\begin{center}
\centerline{\includegraphics[width=0.5\textwidth]{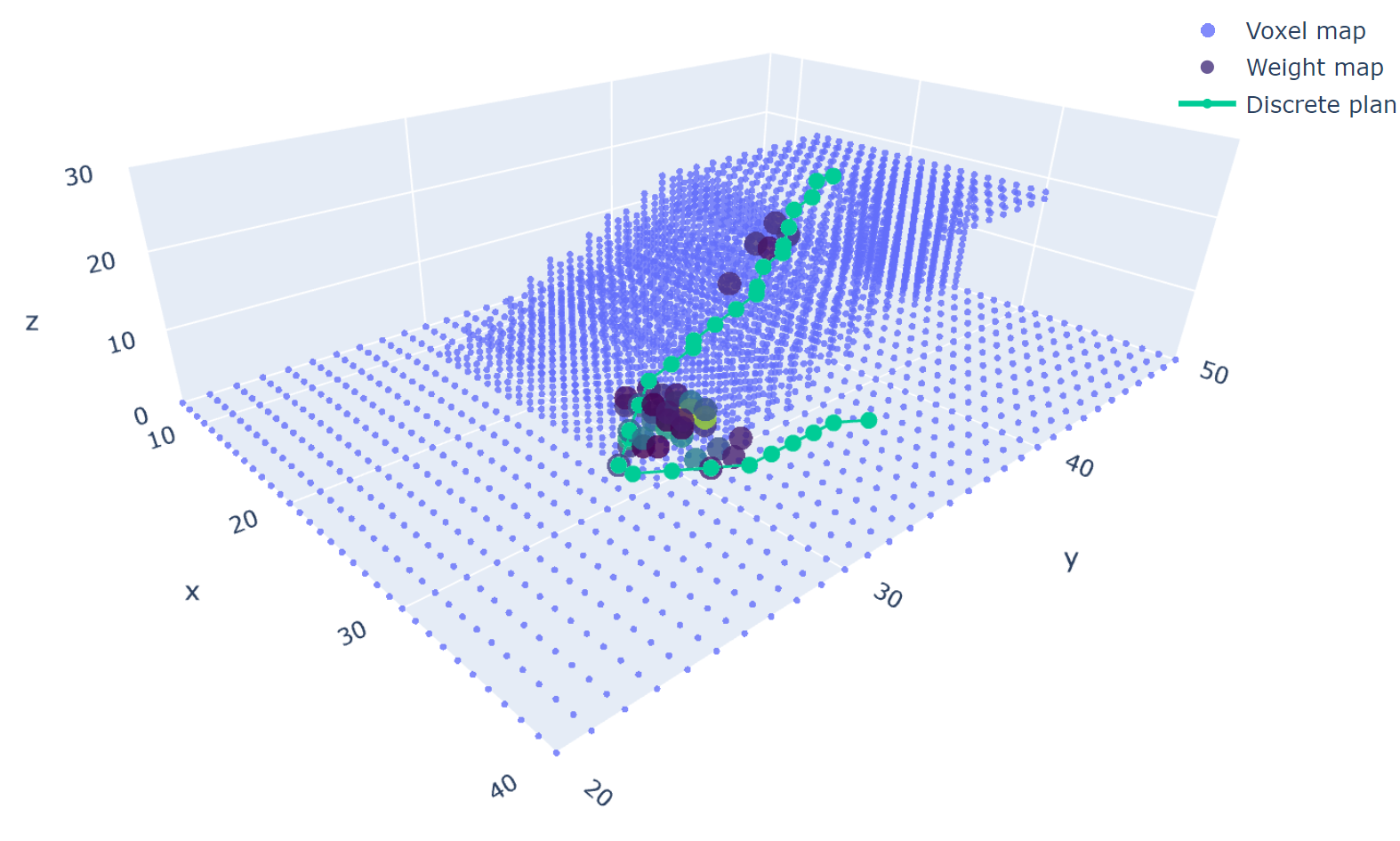}}
\vspace{-0.5em}
\caption{Global path and position weights on the voxel map calculated with autonomous path planning.}
\label{fig:exp1_map}
\end{center}
\end{figure}

The first iteration of planning has the highest computation time and request number. This is correlated with the number of weights that were established. Further iterations have a much less value for parameters described above since the planning algorithm resolves most of the invalid states during its first iteration. 

\subsection{Path Planning with User Guidance}
In the following experiment, the influence of the operator on system performance was evaluated. 
\subsubsection{Procedure} Conditions of the quadruped robot behavior were the same as in the previous experiment. Users were introduced to the VRUI system and practiced setting virtual obstacles several times. Each user then was tasked to set the virtual obstacle rendered as a box object in the VR environment in a way that supports robot navigation at the bottom of the stairs. A new object in the scene was then processed by the Mapping module, which updated the voxel map. 
\subsubsection{Participants} 
Ten participants (three females) aged 22 to 28 years (mean = 24.1) were invited to the experiment. Two of the participants regularly worked with VR interfaces, one had never interacted with VR, and others were interacting with VR systems several times.

\subsubsection{Results} The performance of the planning algorithm is presented in Table~\ref{tab:exp2}.

\begin{table}[htbp]
\caption{Performance of Path Planning Algorithm with User Interaction}
\begin{center}
\vspace{-0.5em}
\begin{tabular}{|c|c|c|c|c|c|c|}
\hline
Iteration number     & 1 & 2 & 3 & 4 & 5 & 6 \\ \hline
\begin{tabular}[c]{@{}c@{}}
Computation\\ time, sec\end{tabular} & 6.6 & 5.8 & 4.9 & 4.5 & 4.4 & 4.5 \\ \hline
\begin{tabular}[c]{@{}c@{}}Path length \\ (number of states)\end{tabular}  & 487 & 472 & 470 & 471 & 470 & 470 \\ \hline
Request count      & 10 & 4 & 5 & 2 & 1 & 1 \\ \hline
\end{tabular}
\label{tab:exp2}
\end{center}
\end{table}

The operator placed a virtual obstacle during the first iteration of the planning algorithm. Dynamics of setting weights are shown in Fig.~\ref{fig:weightsSetted_exp2}.
\begin{figure}[htbp]
\begin{center}
\centerline{\includegraphics[width=0.5\textwidth]{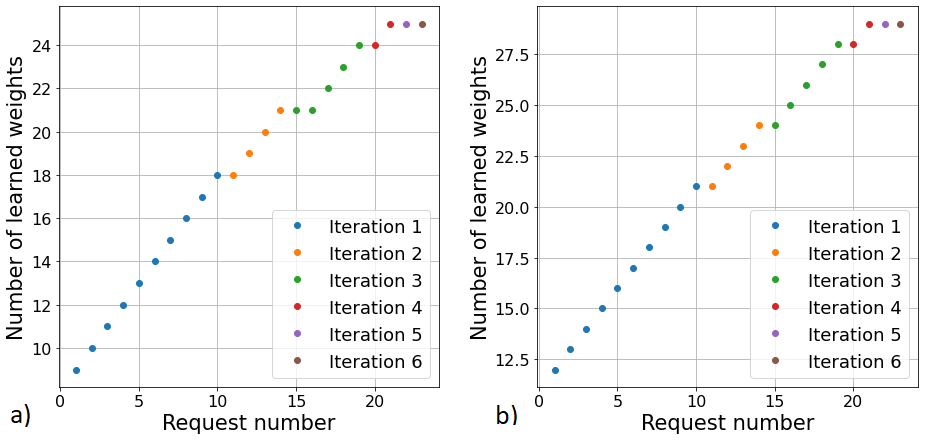}}
\caption{Dynamics of setting (a) position and (b) action weights during planning process with user guidance.}
\label{fig:weightsSetted_exp2}
\end{center}
\end{figure}

The interaction of the user decreased the computational time of the first iteration by 32.3\%. The results revealed that the total time required to find an optimal path was 66.16\% of the same time without user guidance. The view of the global path with learned positional weights is shown in Fig.~\ref{fig:exp2_map}.

\begin{figure}[!h]
\begin{center}
\centerline{\includegraphics[width=0.5\textwidth]{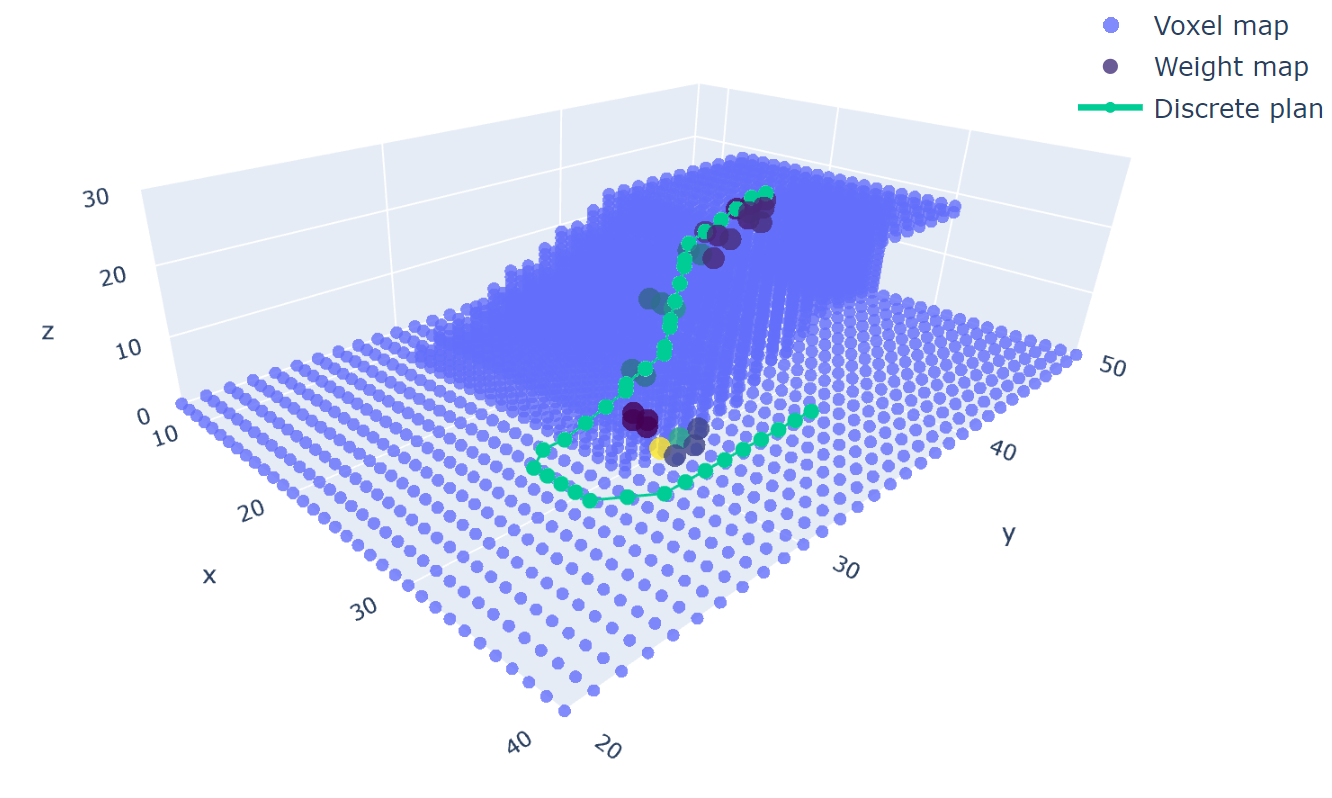}}
\vspace{-0.5em}
\caption{Global path and position weights on the voxel map calculated with user guidance.}
\label{fig:exp2_map}
\end{center}
\end{figure}

In addition, the total number of requests decreased. On the other side, the optimal path without operator influence is shorter by 71 robot states.
The position of the virtual object belongs to the cluster of positional weights acquired from the previous experiment. It reflects on the number of weights that were provided by the local planning algorithm. 

\subsection{User Evaluation of the System by NASA-TLX based Questionnaire}

The additional user study was conducted to validate the performance of the system when operating under user supervision. Participants were asked to assist path planning algorithm. Each user had a short briefing about the virtual interface before the experiment. In addition, they had one attempt to test the interface before the experiment. Participants were not limited in time and number of virtual obstacles to set. In addition, the algorithm was launched one time autonomously to compare results with the user guidance approach. We measured the length of the final constant final path, mean learning duration per iteration, the total number of requests from the local planning algorithm, and the total number of learned positional and action weights. 

We invited ten participants (three females) aged 22 to 28 years (mean = 24.1). Two of the subjects regularly worked with VR interfaces, one had never interacted with VR, and others were familiar with VR systems. The results of the experiment are shown in Fig.~\ref{fig:exp3_stat}. 

\begin{figure}[htbp]
\begin{center}
\centerline{\includegraphics[width=0.5\textwidth]{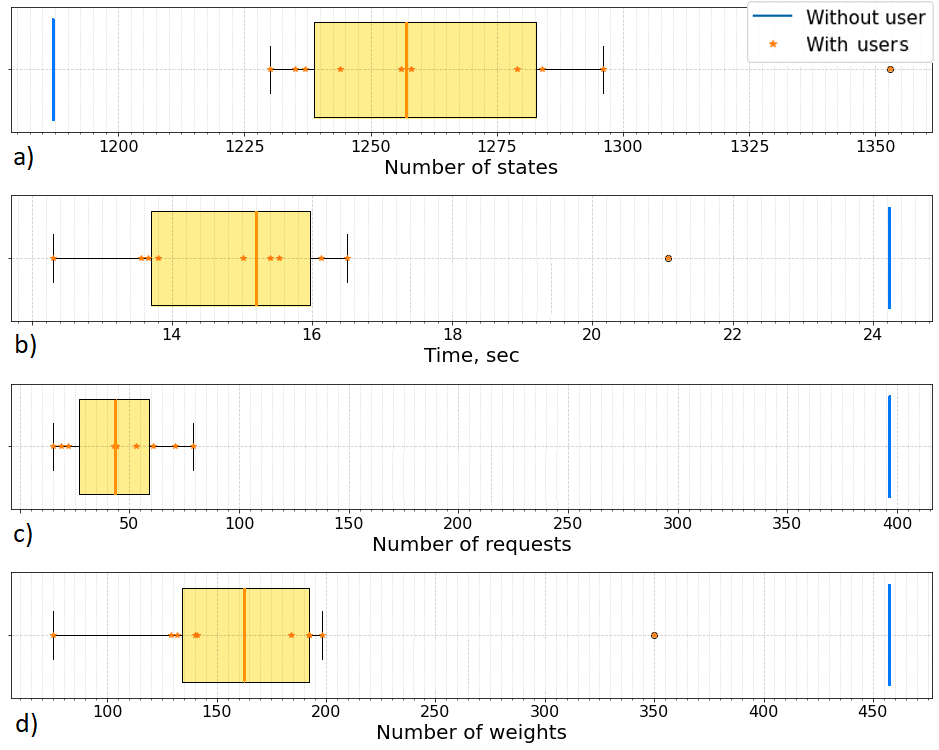}}
\caption{Comparison between supervised and autonomous system performance in: (a) length of the robot path, (b) average time of path generation, (c) total number of request from local to global planner, (d) total number of learned weights. }
\label{fig:exp3_stat}
\end{center}
\vspace{-0.5em}
\end{figure}

The results show that the mean path length is equal to 1266.1 robot states, whereas the path length without interaction is 1187 states (6.66\% more). The mean learning time is 15.623 sec against 24.252 sec without guidance (35.58\% less). The mean total number of requests is 183.14 and the mean number of learned weights is 183.143 against 397 and 314 respectively. However, for the user without experience in VR the path length, training time, and the number of learned weights are out of the confidence interval.

After the experiment, each participant completed a questionnaire based on the NASA Task Load Index (NASA-TLX). The participants provided feedback on six questions:
 
\begin{itemize}
 \item Mental Demand: How much perceptual and mental activity was required for the quadruped platform control? Was this task simple or complex? (Low — High)
 \item Physical Demand: How much physical activity was required? Was the task easy or demanding, slack or strenuous? (Low — High)
 \item Temporal Demand: How much time pressure did you feel due to achieve goal position and orientation during control (Low — High)
 \item Overall Performance: How successful were you in performing the task? (Failure — Perfect)
 \item Effort: Was it hard for you to fulfill the task (mentally and physically) with your level of success? (Low — High)
 \item Frustration Level: Did you feel stressed or annoyed during the task? (Low — High)
\end{itemize}

Results for the NASA-TLX survey are shown in Fig.~\ref{fig:nasa}. 

\begin{figure}[htbp]
 \begin{center}
 \centerline{\includegraphics[width=0.5\textwidth]{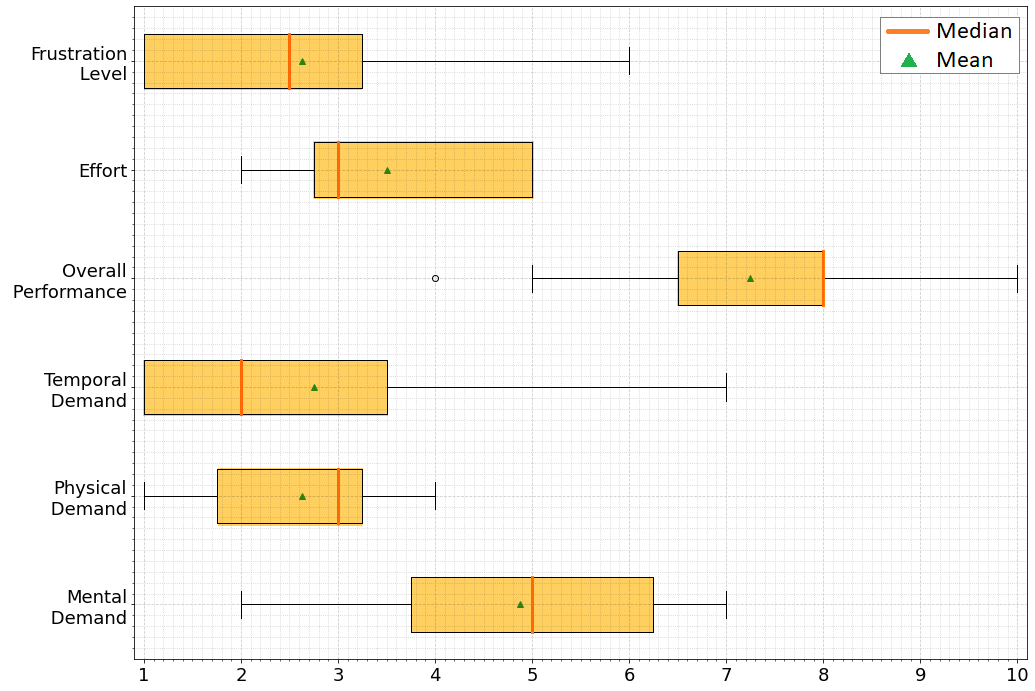}}
 \caption{Subjective feedback on 10-point NASA-TLX based Likert scale.}
 \label{fig:nasa}
 \end{center}
 \vspace{-1 em}
\end{figure}

We conducted a chi-square analysis based on the frequency of answers in each category. The results showed that the experiment parameters are all independent (min $p = 0.12 > 0.05$). In summary, all participants did not feel any additional physical effort during the gesture control performance (mean 2.3 out of 10). The majority of users did not experience time pressure during the experiment (mean 2.9 out of 10), but two participants rated time demand above average. All participants did not feel any stress or frustration (mean 2.5 out of 10).

\section{Conclusion and Future Work}
This paper presents a novel conceptual approach for teaching quadruped robot navigation through user-guided path planning in VR. The HyperGuider framework contains global and local path planners, allowing the robot to adjust its path with few iterations and pass over cluttered and rough terrain. The developed VRUI module allows users to set virtual obstacles in the environment, accelerating path computation. The results of the experiments showed that the assistance of the user increases the speed of the algorithm by 35.58\% on average. This also leads to non-critical elongation of the robot path (of 6.66\% in the tested scenario) in comparison with autonomous planning. In addition, the results of the NASA-TLX survey showed that participants did not feel additional physical effort during the interaction with robots in VR (mean 2.3 out of 10). The majority of users did not experience time pressure during the experiment (mean 2.9 out of 10) and were satisfied with their performance (7.1 out of 10).

Our future work will be divided into two main streams. Firstly, we are going to improve the planning algorithm for quadruped robots by investigating different approaches to setting positional and action weights. Additional investigations can be conducted on the gait planning algorithms and path smoothing process. Secondly, the VR interface is going to be enhanced to allow both generating virtual obstacles and interaction with the robot path by adding feature points. The additional experiment will be conducted to compare different interfaces of human-robot interaction. The developed system is planned to be tested on the real terrain-sensitive quadruped robot DogTouch \cite{Mudalige_2022}.

\section{Acknowledgments}
The reported study was funded by RFBR and CNRS, project number 21-58-15006.

\bibliographystyle{IEEEtran}
\bibliography{bibliography}

\end{document}